\documentclass[11pt,a4paper]{article}
\usepackage[hyperref]{acl2020}
\usepackage{times}
\usepackage{graphicx}
\usepackage{latexsym}
\usepackage{algorithm}
\usepackage{amsmath}
\usepackage[noend]{algpseudocode}

\usepackage{microtype}

\aclfinalcopy % Uncomment this line for the final submission
\def\aclpaperid{***} %  Enter the acl Paper ID here

\title{\textit{How to Grow a (Product) Tree}\\Personalized Category Suggestions for eCommerce Type-Ahead}

\author{Jacopo Tagliabue \\
  Coveo Labs, New York, USA  \\
  \texttt{jtagliabue@coveo.com}\\\And
  Bingqing Yu \\
  Coveo, Montreal, CA \\
  \texttt{cyu2@coveo.com} \\\And
  Marie Beaulieu\\
  Coveo, Quebec, CA \\
  \texttt{mabeaulieu@coveo.com} \\
  }

\date{}

\begin{document}
\maketitle
\begin{abstract}
In an attempt to balance precision and recall in the search page, leading digital shops have been effectively nudging users into select category facets as early as in the type-ahead suggestions.~In~\textit{this} work, we present~\textbf{SessionPath}, a novel neural network model that improves facet suggestions on two counts: first, the model is able to leverage session embeddings to provide scalable personalization; second,~\textbf{SessionPath} predicts facets by \textit{explicitly} producing a probability distribution at each node in the taxonomy path. We benchmark~\textbf{SessionPath} on two partnering shops against count-based and neural models, and show how business requirements and model behavior can be combined in a principled way.
\end{abstract}

\section{Introduction}
\begin{figure}
\begin{center}
    \fbox{\includegraphics[width=6.0cm]{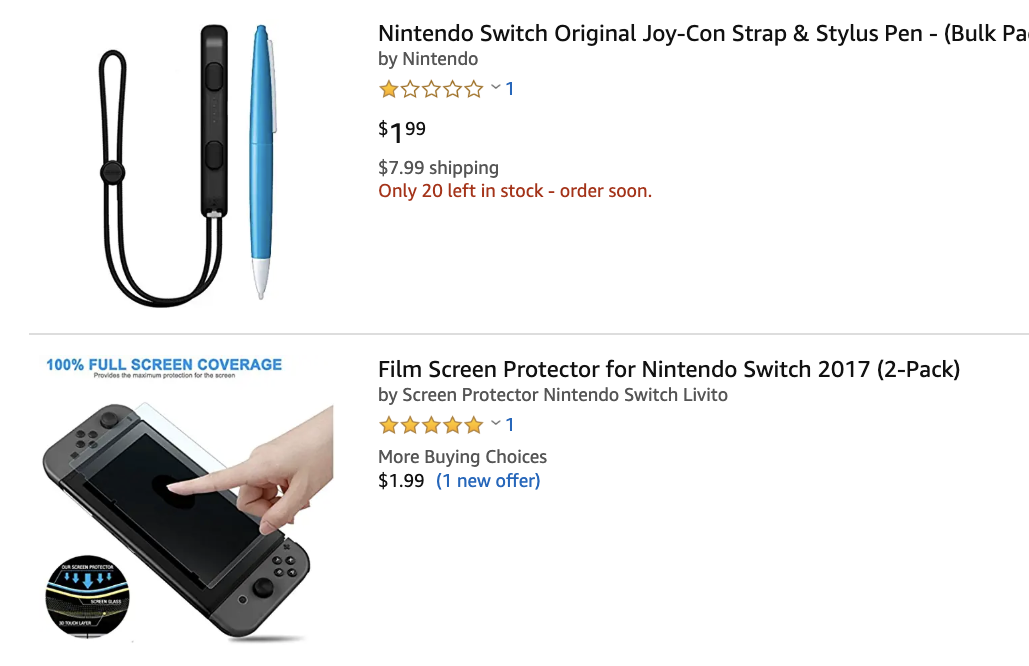}}   
\end{center}
\caption{Price re-ordering on~\textit{Amazon.com}, showing degrading relevance in the result set when querying for a console - ``nintendo switch'' - and then re-ranking based on price. }
\label{fig:amazon}
\end{figure}

Modern eCommerce search engines need to work on millions of products; in an effort to fight ``zero result" pages, digital shops often sacrifice \textit{precision} to increase \textit{recall}\footnote{The ``nintendo switch'' query for a gaming console returns 50k results on~\textit{Amazon.com} at the time of drafting this footnote; 50k results are more products than the entire catalog of a mid-size shop such as~\textbf{Shop 1} below.}, relying on \textit{Learn2Rank}~\citep{10.1561/1500000016} to show the most relevant results in the top positions~\citep{10.1145/1148170.1148246}. While this strategy is effective in web search, when users rarely go after page one \citep{10.1145/1008992.1009079, 10.1145/1240624.1240691}, it is only partially successful in product search: shoppers may spend time browsing several pages in the result set and \textit{re-order} products based on custom criteria (Figure \ref{fig:amazon}); analyzing industry data, up to 20\% of clicked products occur \textit{not} on the first page, with re-ranking in approximately 10\% of search sessions. 

\begin{figure}
\begin{center}
    \fbox{\includegraphics[width=6.0cm]{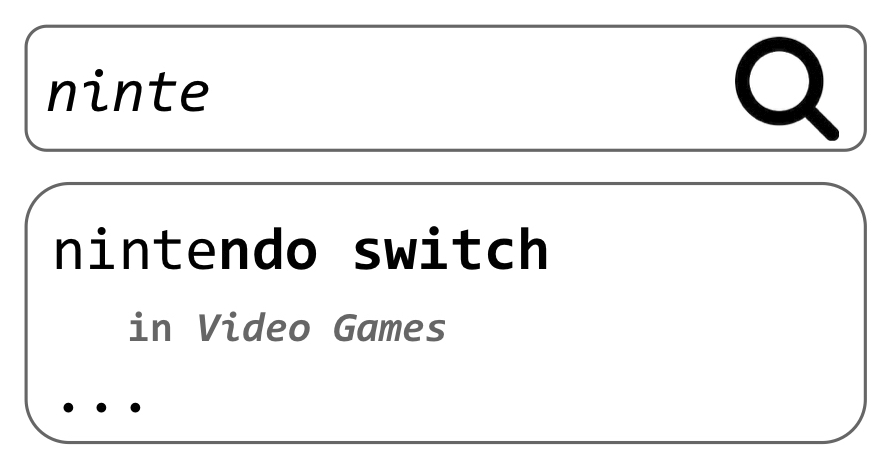}}   
\end{center}
\caption{Facet suggestions during type-ahead: shoppers can be nudged to pick a facet \textit{before} querying, to help the search engine present more relevant results. }
\label{fig:facetSelection}
\end{figure}

Leading eCommerce websites leverage machine learning to suggest \textit{facets} - i.e. product categories, such as \textit{Video Games} for ``nintento switch'' - \textit{during} type-ahead (Figure~\ref{fig:facetSelection}): narrowing down candidate products explicitly by matching the selected categories, shops are able to present less noisy result pages and increase the perceived relevance of their search engine. In~\textit{this} work we present \textbf{SessionPath}, a scalable and personalized model to solve facet prediction for type-ahead suggestions: given a shopping session and candidate queries in the suggestion dropdown menu, the model is asked to predict the best category facet to help users narrow down search intent. A big advantage of~\textbf{SessionPath} is that it can complement any existing stack by adding facet prediction to items as retrieved by the type-ahead API. 

We summarize the main contributions of \textit{this} work as follows:

\begin{itemize}
    \item we devise, implement and benchmark several models of incremental complexity (as measured by features and engineering requirements); starting from a non-personalized count-based baseline, we arrive at  \textbf{SessionPath}, an encoder-decoder architecture that explicitly models the real-time generation of paths in the catalog taxonomy;
    \item we discuss the importance of false positives and false negatives in the relevant business context, and provide decision criteria to adjust the precision/recall boundary after training. By combining the predictions of the neural network with a \textit{decision module}, we show how model behavior can be tuned in a principled way by human decision makers, without interfering with the underlying inference process or introducing \textit{ad hoc} manual rules. 
\end{itemize}

To the best of our knowledge, \textbf{SessionPath} is the first type-ahead model that allows \textit{dynamic} facet predictions: linguistic input and in-session intent are combined to adjust the target taxonomy depth (\textit{sport / basketball} vs \textit{sport / basketball / lebron}) based on real-time shopper behavior and model confidence. For this reason, we believe the methods and results here presented will be of great interest to any digital shop struggling to strike the right balance between precision and recall in a catalog with tens-of-thousands-to-millions of items.

\section{Less (Choice) is More: Considerations From Industry Use Cases}
The problem of narrowing down the result set before re-ranking is a known concern for mid-to-big-size shops: as shown in Figure~\ref{fig:amazon}-A, a common solution is to invite shoppers to select a category facet when still typing. Aside from UX considerations, restricting the result set may be beneficial for other reasons. On one hand, decision science proved that providing shoppers with \textit{more} alternatives is actually less efficient (the so-called "paradox of choice" \citep{10.1086/651235, articleChoice2001}) - insofar as \textbf{SessionPath} helps avoiding unnecessary ``cognitive load'', it may be a welcomed ally in fighting irrational decision making; on the other, by restricting result set through facet selection, the model may reduce the long-tail effect of many queries on product visibility: when results are too many and items frequently changed, standard \textit{Learn2Rank} approaches tend to penalize less popular items \citep{inproceedingsAbdollahpouri, 10.5555/1197299}, which end up buried among noisy results far from the first few pages and never collect enough relevance feedback to rise through the top.

\begin{figure}
\begin{center}
    \includegraphics[width=7.0cm]{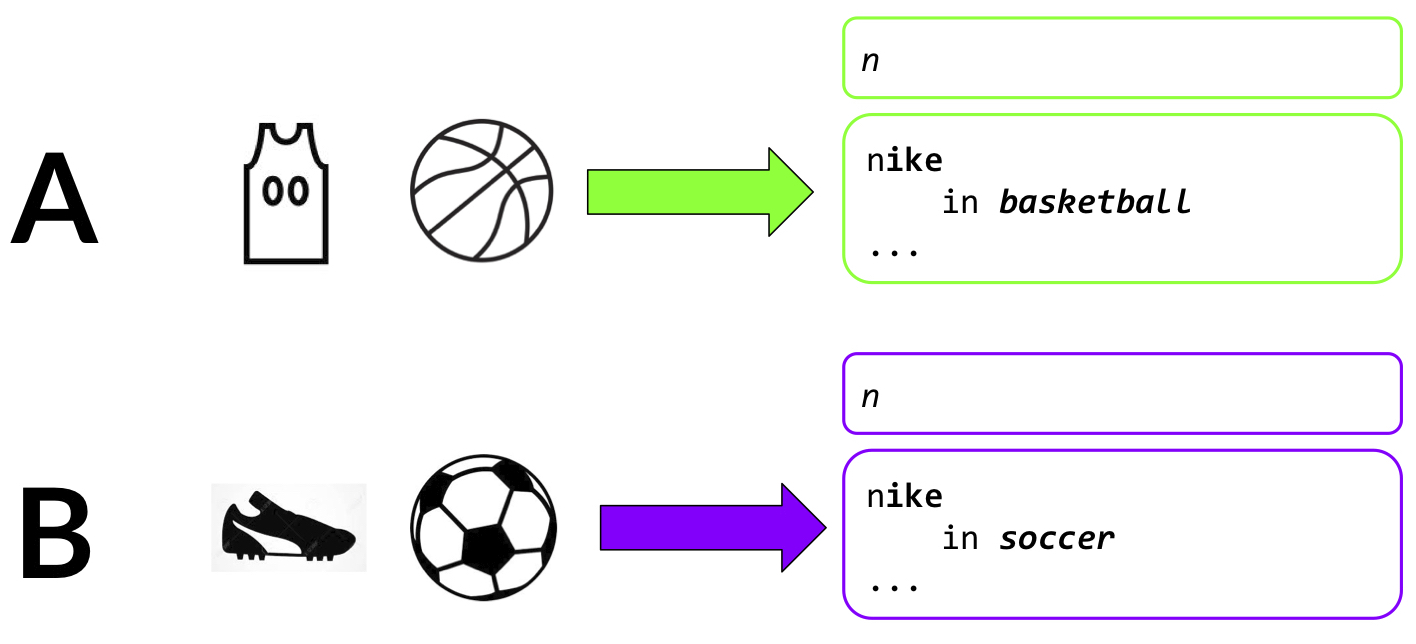}
\end{center}
\caption{Shoppers in \textbf{Session A} and \textbf{Session B} have different sport intent, as shown by the products visited. By combining linguistic and behavioral in-session data, \textbf{SessionPath} provides in real-time personalized facet suggestions to the same ``nike'' query in the type-ahead.}
\label{fig:usecase}
\end{figure}

In \textit{this} work, we extend industry best practices of facet suggestion in type-ahead by providing a solution that is dynamic in two ways: i) given the same query, session context may be used to provide a contextualized suggestion (Figure~\ref{fig:usecase}); ii) given two queries, the model will decide in real-time how deep in the taxonomy path the proposed suggestion needs to be (Figure~\ref{fig:functional}): for some queries, a generic facet may be optimal (as we do not want to narrow the result set \textit{too much}), for others a more specific suggestion may be more suitable. Given the natural trade-off between \textit{precision} and \textit{recall} at different depths, Section~\ref{sec:decision} is devoted to provide a principled solution.

\section{Related Work}

\textbf{Facet selection}. Facet selection is linked to \textit{query classification} on the research side \citep{8622008,Skinner2019EcommerceQC} and \textit{query scoping} on the product side, i.e. pre-selecting, say, the facet \textit{color} with value \textit{black} for a query such as ``black basketball shoes''~\citep{10.1016/j.scico.2013.07.019, 10.1145/2505515.2505664}. Scoping may result in an aggressive restriction of the result set, lowering \textit{recall} excessively: in most cases, an acceptable shopping experience would need to combine scoping with \textit{query expansion}~\citep{Diaz2016QueryEW}. \textbf{SessionPath} is more flexible than query classification, by supporting explicit path prediction and incorporating in-session information; it is more gentle than scoping (by nudging transparently the final user instead of forcing a selection behind the scene); it is more principled than expansion in balancing precision and recall.

\textbf{Deep Learning in IR}. The development of deep learning models for IR has been mostly restricted to the retrieve-and-rerank paradigm~\citep{Mitra2017NeuralMF, Guo2016ADR}. Some recent works have been focused specifically on ranking suggestions for type-ahead: neural language models are proposed by~\citet{inproceedingsparj2017, conf/sigir/WangZMDK18}; specifically in eCommerce, \citet{DBLP:journals/corr/abs-1905-01386} employs \textit{fastText} to represent queries in the ranking phase and \citet{CoveoECNLP20} leverages deep image features for in-session personalization. While \textit{this} work employs deep neural networks both for feature encoding and the inference itself, the proposed methods are agnostic on the underlying retrieval algorithm, as long as platforms can enrich type-ahead response with the predicted category. By providing a gentle entry point into existing workflows, a great product strength of \textbf{SessionPath} is the possibility of deploying the new functionalities with minimal changes to any architecture, neural or traditional (see also Appendix~\ref{sec:architecture}).

\section{Problem Statement}
\label{sec:statement}
Suggesting a category facet can be modelled with the help of few formal definitions. A target shop $E$ has products $p_1, p_2, ... p_n \in P$ (e.g. \textit{nike air max 97}) and categories $c_{1,1}, c_{1,2}, ... c_{n,m} \in C$, where $c_{n,m}$ is the category $n$ at \textit{depth} $m$ (e.g. at $m=1$, [\textit{soccer, volley, football, basketball}], at $m=2$ [\textit{shoes, pants, t-shirts}], etc.); a taxonomy tree $T_m$ is an \textit{indexed} mapping $P \mapsto C_{m}$, assigning a category to products for each depth $m$ (e.g.~\textit{air max 97}~$\mapsto_{0}$~\textit{root},$\mapsto_{1}$~\textit{soccer},~$\mapsto_{2}$~\textit{shoes},~$\mapsto_{3}$~\textit{messi}~etc.);~\textit{root} is the base category in the taxonomy, and it is common to all products (we will omit it for brevity in all our examples). In what follows, we use~\textit{path} to denote a sequence of categories (hierarchically structured) in our target shop (e.g.~\textit{root / soccer / shoes / messi}), and \textit{nodes} to denote the categories in a path (e.g.~\textit{soccer} is a node of ~\textit{soccer / shoes / messi}).

Given a browsing session $s$ containing products $p_x, p_y, ... p_z$, and a candidate type-ahead query $q$, the model's goal is to learn both the optimal depth value $m$ and, for each $k \leq m$, a contextual function $f(q, s) \mapsto C_k$. As we shall see in the ensuing section, \textbf{SessionPath} solution to this challenge is two-fold: a model generating a path first, and a decision module to pick the appropriate depth $m$ (Figure~\ref{fig:functional}).

\section{Baseline and Personalized Models}
\label{sec:SessionPathexplanation}
\begin{figure}
\begin{center}
    \includegraphics[width=7.5cm]{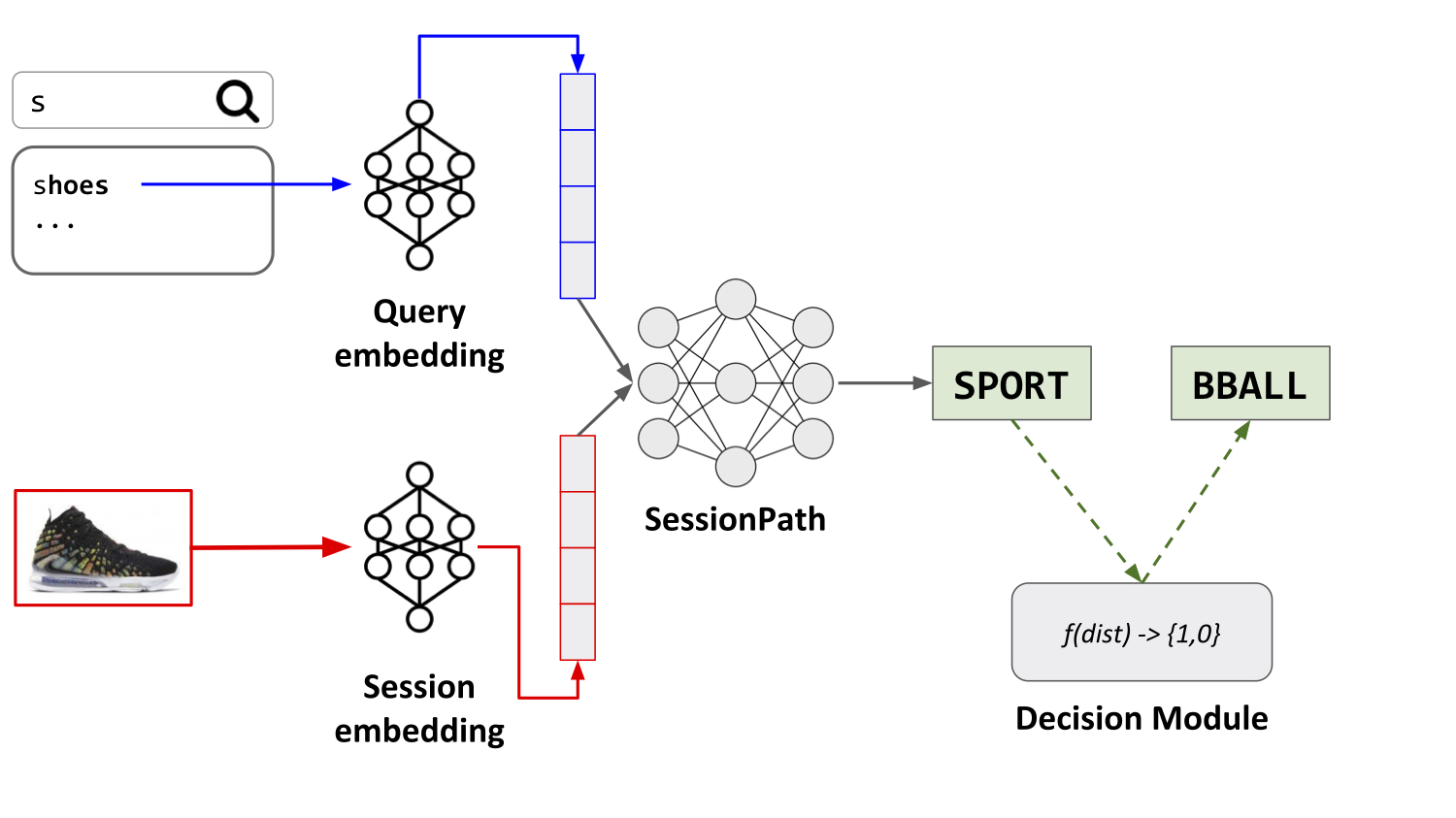}    
\end{center}
\caption{Functional flow for~\textbf{SessionPath}: the current session and the candidate query ``shoes'' are embedded and fed to the model; the distribution over possible categories at each step of the taxonomy is passed to a \textit{decision module}, that either cuts the generation at that step or includes the step in the final prediction. The decision process is repeated until either the module cuts or a max-length path is generated.}
\label{fig:functional}
\end{figure}

We approach the challenge \textit{incrementally}, by first developing a count-based model (\textbf{CM}) that learns a mapping from queries to all paths (i.e. \textit{sport} and \textit{sport / soccer} are treated purely as ``labels", so they are two completely unrelated target classes for the model); \textbf{CM} will both serve as a baseline for more sophisticated methods and as a fast reference implementation not requiring any deep learning infrastructure. We improve on \textbf{CM} with \textbf{SessionPath}, a model based on deep neural networks. From a product perspective, it is important to remember (Figure~\ref{fig:functional}) that a \textit{decision module} is called \textit{after} a path prediction is made: we discuss how to tune this crucial part after establishing the general performance of the proposed models.

\subsection{A Baseline Model}
The intuition behind the count-based model is that we may gain insights on relevant paths linked to a query from the clicks on search results. Therefore, we can build a simple count-based model by creating a map from each query in the search logs to their frequently associated paths. To build this map, we first retrieve all products clicked after each query, along with their path; for a given query, we can then calculate the percentage of occurrence of each path in the clicked products. Since the model is not hierarchical, it is important to note that \textit{sport} and \textit{sport / basketball} will be treated as completely disjoint target classes for the prediction. To avoid noisy results, we empirically determined a frequency threshold for paths to be counted as relevant to a certain query (80\%); at prediction time, given a query in the training set, we retrieve all the paths associated with it and return the one with longest depth; for unseen queries, no prediction is made. 

\subsection{Modelling Session Context and Taxonomy Paths}
\label{model:dl}
The main conceptual improvements over \textbf{CM} are three:

\begin{itemize}
    \item \textbf{SessionPath} produces predictions also for queries not in the training set;
    \item \textbf{SessionPath} introduces personalization, by combining the linguistic information contained in the query with in-session shopping intent;
    \item \textbf{SessionPath} is trained to produce the most accurate path by explicitly making a new prediction \textit{at each node}, not predicting paths in a \textit{one-out-of-many} scenario; in other words,~\textbf{SessionPath} knows that~\textit{sport} and \textit{sport / basketball} are related, and that the second path is generated from the first when a given distribution over sport activities is present.
\end{itemize}

To represent the current session in a dense architecture, we first train a skip-gram \textit{prod2vec} model over user data for the entire website, mapping product to 50-dimensional vectors~\citep{Mikolov:2013, Grbovic15}.~At training and serving time~\textbf{SessionPath} retrieves the embeddings of the products in the target session, and use average pooling to calculate the context vector from the sequence of embeddings, as shown by \citet{10.1145/2959100.2959190, CoveoECNLP20}. To represent the candidate query, an encoding of linguistic behavior that generalizes to unseen queries is needed. We tested different strategies:

\begin{itemize}
    \item \textit{word2vec}:~we train a skip-gram model from~\citet{NIPS2013_5021} over product short descriptions from the catalog. Since most search queries are less than three words long, we opted for a simple and fast average pooling of the embeddings in the tokenized query;
    \item \textit{character-based language model}: inspired by \citet{Skinner2018ProductCW}, we train a char-based language model (single LSTM layer with hidden dimension $50$) on search logs and product descriptions from the target shop; a standard LSTM approach was found ineffective in preliminary tests, so we opted instead for using the ``Balanced pooling'' strategy from \citet{Skinner2019EcommerceQC}, where the dense representation for the query is obtained by taking the last network state and then concatenating it together with average-pooling~\cite{wang-etal-2018-lrmm}, max-pooling, and min-pooling;
    \item \textit{pre-trained language model}: we map the query to a 768-size vector using BERT~\citep{devlin-etal-2019-bert} (as pre-trained for the target language by~\citet{magnini2006annotazione});
    \item \textit{Search2Prod2Vec}~+~\textit{unigrams}: we propose a ``small-data'' variation to \textit{Search2Vec} by~\citet{10.1145/2911451.2911538}, where queries (on a web search engine) are embedded through events happening before and after the search event. Adapting the intuition to product search, we propose to represent queries through the embeddings of products clicked in the search result page; in particular, each query $q$ is the weighted average of the corresponding \textit{prod2vec} embeddings; it can be argued that the \textit{clicking} event is analogous to a ``pointing'' signal~\citep{Tagliabue2019LexicalLA}, when the~\textit{meaning} of a word (``shoes'') is understood as a function from the string to a set of objects falling under that concept (e.g.~\citet{10.5555/335289}). In the spirit of compositional semantics~\citep{Baroni2014FregeIS}, we generalize this representation to unseen queries by building a unigram-based language model, so that ``nike shoes'' gets its meaning from the composition (average pooling) of the meaning of \textit{nike} and \textit{shoes}.
\end{itemize}

To generate a path explicitly, we opted for an encoder-decoder architecture. The encoder employs the wide-and-deep approach popularized by \citet{Cheng2016WideD}, and concatenates textual and non-textual feature to obtain a wide representation of the current context, which is passed through a dense layer to represent the final encoded state. The decoder is a word-based language model~\citep{Zoph2016NeuralAS} which produces a sequence of nodes (e.g. \textit{sport}, \textit{basketball}, etc.) conditioned on the representation created by the encoder; more specifically, the architecture of the decoder consists of a single LSTM with 128 cells, a fully-connected layer and a final layer with softmax output activation. The output dimension corresponds to the total number of distinct nodes found in all the paths of the training data, including two additional tokens to encode the start-of-sequence and end-of-sequence. For training, the decoder uses the encoded information to fill its initial cell states; at each timestep, we use teacher forcing to pass the target character, offset by one position, as the next input character to the decoder~\cite{Williams89alearning}. Empirically, we found that robust parameters for the deep learning methods are a learning rate of $0.001$, time decay of $0.00001$, early stopping with $patience=20$, and mini-batch of size $128$; furthermore, the Adam optimizer with cross-entropy loss is used for all networks, with training up to 300 epochs. Once trained, the model can generate a path given an encoded session representation and a start-of-sequence token: after the first step, the decoder uses autoregression sequence generation~\cite{Bahdanau2015NeuralMT} to predict the next output token.

\section{Dataset}

\begin{table}
\centering
\begin{tabular}{lcc}
\hline \textbf{Shop} & \textbf{Queries (with context)} & \textbf{Products} \\ \hline
Shop 1 & 270K (185K) & 29.699 \\
Shop 2 & 270K (227K) & 93.967 \\
\hline
\end{tabular}
\caption{\label{tbl:dataset} Descriptive statistics for the dataset. }
\end{table}

We leverage behavioral and search data from two partnering shops in~\textit{Company}'s network: \textbf{Shop 1} and \textbf{Shop 2} have uniform data ingestion, making it easy to compare how well models generalize; they are mid-size shops, with annual revenues between 20 and 100 million dollars. \textbf{Shop 1} and \textbf{Shop 2} differ however in many respects: they are in different verticals (\textit{apparel} vs \textit{home improvement}), they have a different catalog structure (603 paths organized in 2-to-4 levels for each product vs 985 paths in $3$ levels for all products), and different traffic (top 200k vs top 15k in the Alexa Ranking). Descriptive statistics for the training dataset can be found in Table~\ref{tbl:dataset}: data is sampled for both shops from June-August in 2019; for testing purposes, a completely disjoint dataset is created using events from the month of September.

\section{Experiments}
\label{sec:experiments}
We perform offline experiments using search logs for \textbf{Shop 1} and \textbf{Shop 2}: for each search event in the dataset, we use products seen before the query (if any) to build a session vector as explained in Section \ref{model:dl}; the path of the products \textit{clicked after} the query is used as the target variable for the model under examination.

\begin{table}
\centering
\begin{tabular}{lccc}
\hline\textbf{Model} & \textbf{D=1} & \textbf{D=2} & \textbf{D=last} \\ \hline
CM & 0.63 & 0.53 & 0.22\\
MLP+BERT & 0.72 & 0.59 & 0.33\\
SP+BERT & 0.77 & 0.64 & 0.40 \\
SP+LSTM & 0.79 & 0.68 & 0.43\\
SP+W2V & 0.82 & 0.71 & 0.46 \\
SP+SV & \textbf{0.87} & \textbf{0.79(0.01)} & \textbf{0.55}\\
\hline
CM & 0.41 & 0.34 & 0.24\\
MLP+BERT & 0.61 & 0.50 & 0.39\\
SP+BERT & 0.66 & 0.55 & 0.45\\
SP+LSTM & 0.67 & 0.57 & 0.46\\
SP+W2V & 0.69 & 0.59 & 0.47 \\
SP+SV & \textbf{0.80} & \textbf{0.71}& \textbf{0.59}\\
\hline
\end{tabular}
\caption{\label{tbl:results1} Accuracy scores for $depth=1$, $depth=2$, $depth=last$, divided by \textbf{Shop 1} (\textit{top}) and \textbf{Shop 2} (\textit{bottom}). We report the mean over $5$ runs, with SD if $SD\geq0.01$.}
\end{table}

\subsection{Making predictions}
We benchmark \textbf{CM} and \textbf{SessionPath} from Section~\ref{sec:SessionPathexplanation}, plus a multi-layer perceptron (\textbf{MLP}) to investigate the performance of an intermediate model: while not as straightforward as \textbf{CM}, \textbf{MLP} is considerably easier to train and serve than \textbf{SessionPath} and it may therefore be a compelling architectural choice for many shops (see Appendix~\ref{sec:architecture} for practical engineering details); \textbf{MLP} concatenates the session vector with the~\textit{BERT} encoding of the candidate query, and produces a distribution over all possible \textit{full-length} paths (\textit{one-out-of-many} classification, where the target class comprises all the paths at the maximum depth for the catalog at hand).~Table~\ref{tbl:results1} shows accuracy scores for three different depth levels in the predicted path:~\textbf{SP+BERT} is \textbf{SessionPath} using~\textit{BERT} to encode linguistic behavior,~\textbf{SP+W2V} is using \textit{word2vec},~\textbf{SP+SV} is using~\textit{Search2Prod2Vec} and ~\textbf{SP+LSTM} is using the language model. Every~\textbf{SessionPath} variant outperforms the count-based and neural baselines, with~\textit{Search2Prod2Vec} providing up to 150\% increase over \textbf{CM} and 67\% over \textbf{MLP}.~\textbf{CM} score is penalized not only by the inability to generalize to unseen queries: even when considering previously seen queries in the test set,~\textbf{SP+SV}'s accuracy is significantly higher ($0.58$  vs $0.27$ at $D=last$), showing that neural methods are more effective in capturing the underlying dynamics. 
Linguistic representations learned \textit{directly} over the target shop outperform bigger models pre-trained on generic text sources, highlighting some differences between general-purpose embeddings and shop-specific ones, and suggesting that off-the-shelf NLP models may not be readily applied to short, keyword-based queries. While fairly accurate,~\textbf{SP+W2V} is much slower to train compared to~\textbf{SP+SV} and harder to scale across clients, as it relies on having enough content in the catalog to train models that successfully deal with shop lingo. On a final language-related note, it is worth stressing that click-based embeddings built for~\textbf{SP+SV} show not just better performance over~\textit{seen} queries (which is expected), but better generalization ability in the \textit{unseen} part as well compared to \textit{BERT} embeddings ($0.82$ vs $0.70$ at $D=1$ for \textbf{Shop 1}, $0.76$ vs $0.63$ for \textbf{Shop 2}).

In the spirit of ablation studies, we re-run~\textbf{SP+SV} and~\textbf{SP+BERT}~\textit{without} session vector. Interestingly enough, context seems to play a slightly different role in the two shops and the two models:~\textbf{SP+BERT} is greatly helped by contextual information, especially for \textit{unseen} queries ($0.28$ vs $0.21$ at $D=last$ for \textbf{Shop 1}, $0.40$ vs $0.15$ for \textbf{Shop 2}), but the effect for~\textbf{SP+SV} is smaller ($0.50$ vs $0.42$ for \textbf{Shop 2}); while models on \textbf{Shop 2} show a bigger drop in performance when removing session information, generally (and unsurprisingly) session-aware models provide better generalization on \textit{unseen} queries across the board. By comparing \textbf{SessionPath} with a simpler neural model (such as~\textbf{MLP}), it is clear that session plays a bigger role in \textit{MLP}, suggesting that \textbf{SessionPath} architecture is able to better leverage linguistic information across cases. 

Finally, we investigate sample efficiency of chosen methods by training on smaller fractions of the original training dataset: Table~\ref{tbl:efficiencyresults} reports accuracy of four methods when downsampling the training set for \textbf{Shop 1} to $1/10^{th}$ and $1/4^{th}$ of the dataset size. \textbf{CM}'s inability to generalize cripples its total score; \textbf{MLP} is confirmed to be simple yet effective, performing significantly better than the count-based baseline; \textbf{SP+SV} is confirmed to be the best performing model, and even with only $1/10^{th}$ of samples outperforms all other models from Table~\ref{tbl:results1}: by leveraging the bias encoded in the hierarchical structure of the products,~\textbf{SP+SV} allows paths that share nodes (\textit{sport}, \textit{sport / basketball}) to also share statistical evidence, resulting in a very efficient learning.

\begin{table}
\centering
\begin{tabular}{lccc}
\hline\textbf{Model (D=last)} & \textbf{1/10} & \textbf{1/4} \\ \hline
CM & 0.18 & 0.20 \\
MLP+BERT & 0.28 & 0.30 \\
SP+BERT & 0.31 & 0.34\\
SP+SV & \textbf{0.51} & \textbf{0.53}\\
\hline
\end{tabular}
\caption{Accuracy scores (\textbf{D=last}) when training on portions of the original dataset for \textbf{Shop 1}.}
\label{tbl:efficiencyresults}
\end{table}

Accuracy provides a strong argument on the efficacy of the proposed models in industry, and it is in fact widely employed in the relevant literature: \citet{10.1145/2505515.2505664} employs click-based accuracy for label prediction, while~\citet{Molino2018COTAIT} (in a customer service use cases) uses accuracy at different depths for sequential predictions that are somewhat similar to \textbf{SessionPath}. However,~\textit{accuracy} by itself falls short to tell the whole story on product decisions: working with \textit{Coveo}'s clients, it is clear that not all shops are born equal - some (e.g. mono-brand fashion shops) strongly favor a smaller and cleaner result page; others (e.g. marketplaces) favor bigger, even if noisier, result sets. Section~\ref{sec:decision} presents our contribution in analyzing the business context and proposes viable solutions.

\subsection{Tuning the decision module}
\label{sec:decision}
\begin{figure}
\begin{center}
    \includegraphics[width=7.0cm]{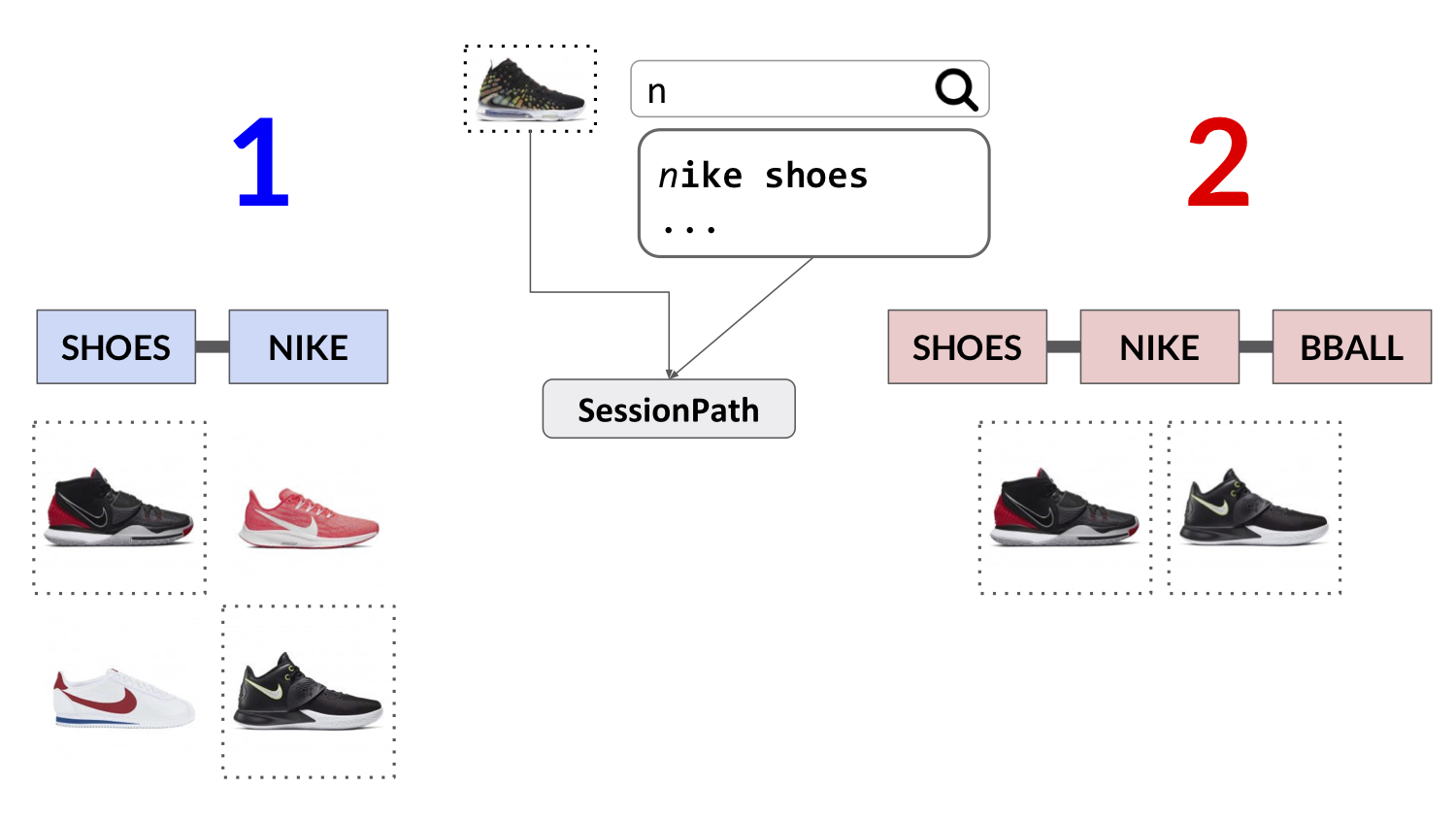}
\end{center}
\caption{Two scenarios for the decision module after \textbf{SessionPath} generates the \textit{shoes / nike / basketball} path, with input query ``nike shoes'' and  Lebron James basketball shoes in the session. In \textit{Scenario 1 (blue)}, we cut the result set after the second node -  \textit{shoes / nike} - resulting in a mix set of shoes; in \textit{Scenario 2 (red)}, we use the full path - \textit{shoes / nike / basketball} - resulting in only basketball shoes (dotted line products). How can we define what is the optimal choice?}
\label{fig:generation}
\end{figure}

Consider the two possible decisions in the scenario depicted in Figure~\ref{fig:generation}: given ``nike shoes'' as query and basketball shoes as session context,~\textbf{SessionPath} prediction is \textit{shoes / nike / basketball}. According to scenario \textbf{1}, a decision is made to cut the path at \textit{shoes / nike}: the resulting set of products contain a mixed set of shoes from the target brand, with no specific sport affinity; in scenario \textbf{2}, the decision module allows the prediction of a longer path, \textit{shoes / nike / basketball}: the result page is smaller and only contains basketball shoes of the target brand. Intuitively, a perfect model would choose \textbf{2} only when it is ``confident'' of the underlying intention, as expressed through the combination of language and behavioral clues; when the model is less confident, it should stick to \textbf{1} to avoid hiding from the shopper's possible interesting products.

To quantify how much confident the model is at any given node in the predicted path, at each node $s_n$ we output the multinomial distribution $d$ over the next node $s_{n+1}$\footnote{Non-existent paths account for less then 0.005\% of all the paths in the test set, proving that \textbf{SessionPath} is able to accurately learn transitions between nodes and suggesting that an explicit check at decision time is unnecessary. Of course, if needed, a ``safety check'' may be performed at query time by the search engine, to verify that filtering by the suggested path will result in a non-empty set.} and calculate the Gini coefficient of $d$, $g(d)$:

\begin{equation}
\tag{GI}
g(d)=\frac{\sum_{i=1}^{n} \sum_{j=1}^{n}\left|x_{i}-x_{j}\right|}{2 n^{2} \bar{x}}
\end{equation}

where $n$ is the total number of classes in the distribution $d$, $x_{i}$ is the probability of the class $i$ and $\bar{x}$ is the mean probability of the distribution.

Once $GI=g(d)$ is computed, a \textit{decision rule} $DR(GI)$ for the decision module in Figure~\ref{fig:functional} is given by:

\[
  DR(x) =
  \begin{cases}
    1 & \text{if $x \geq ct$} \\
    0 & \text{otherwise}
  \end{cases}
\]

where $1$ means that the module is confident enough to add the node to the final path that will be shown to the user, while $0$ means the path generation is stopped at the current node. $ct$ is our \textit{confidence threshold}: since different values of $ct$ imply more or less aggressive behavior from the model, it is important to tune $ct$ by taking into account the relevant business constraints.

Due to the contextual and interactive nature of \textbf{SessionPath}, we turn search logs into a ``simulation'' of the interactions between hypothetical shoppers and our model~\citep{10.1145/3331184.3331362}. In particular, for any given search event in the test dataset - comprising products seen in the session, the actual query issued, all the products returned by the search engine, the products clicked from the shopper in the result page -, and a model prediction (e.g.~\textit{sport / basketball}), we construct two items:

\begin{itemize}
    \item \textbf{golden truth set}: which is the set of the paths corresponding to the items the shopper deemed relevant in that context (relevance is therefore assessed as \textit{pseudo-feedback} from clicks);
    \item \textbf{filtered result set}: which is the set of products returned by the engine,~\textit{filtering} out those not matching the prediction by the model (i.e. simulating the engine is actually working with the categories suggested by \textbf{SessionPath}).
\end{itemize}

With the \textit{golden truth set}, the \textit{filtered result set} and the original \textit{result page}, we can calculate \textit{precision} and \textit{recall} at different values of $ct$ (please refer to Appendix~\ref{sec:metricexample} for a full worked out example).

\begin{table}
\centering
\begin{tabular}{ccc}
\hline\textbf{Gini Threshold} & \textbf{Precision} & \textbf{Recall}\\ \hline
0.996 & 0.65 & 0.99\\
0.993 & 0.82 & 0.91\\
0.990 & 0.93 & 0.77\\
0.980 & 0.99 & 0.74\\
\hline
\end{tabular}
\caption{Precision and recall at different decision thresholds for \textbf{Shop 1}.}
\label{tbl:gini}
\end{table}

Table~\ref{tbl:gini} reports the chosen metrics calculated for \textbf{Shop 1} at different values of $ct$; the trade-off between the two dimensions makes all the point Pareto-optimal: there is no way to increase performance in one dimension without hurting the other. Going from the first configuration ($ct=0.996$) to the second ($ct=0.993$) causes a big jump in the metric space, with the model losing some recall to gain \textit{considerably} in precision. 
%Figure~\ref{fig:pareto} depicts the case for~\textit{unseen} queries, by plotting in a two-dimensional space the metrics obtained on predictions for queries never seen in the training set. 
To get a sense of how the model is performing in practice, Figure~\ref{fig:predictions} shows three sessions for the query ``nike shoes'': when session context is empty (session $1$), the model defaults to the broadest category (\textit{sneakers}); when session is \textit{running}-based or \textit{basketball}-based, the model adjusts its aggressiveness depending on the threshold we set. It is interesting to note that while the prediction for $2$ at $ct=0.97$ is wrong at the last node (product is \textit{a7}, not \textit{a3}), the model is still making a \textit{reasonable} guess (e.g. by guessing sport and brand correctly).

In our experience, the adoption of data-driven models in traditional digital shops is often received with some skepticism over the  ``supervision'' by business experts~\citep{Baer2017}: a common solution is to avoid the use of neural networks, in favor of model interpretability.~\textbf{SessionPath}'s decision-based approach dares to dream a different dream, as the proposed architecture shows that we can retain the accuracy of deep learning and still provide a meaningful interface to business users -- here, in the form of a precision/recall space to be explored with an easy-to-understand parameter.

\section{Conclusions and Future Work}
\begin{figure}
\begin{center}
    \includegraphics[width=7.7cm]{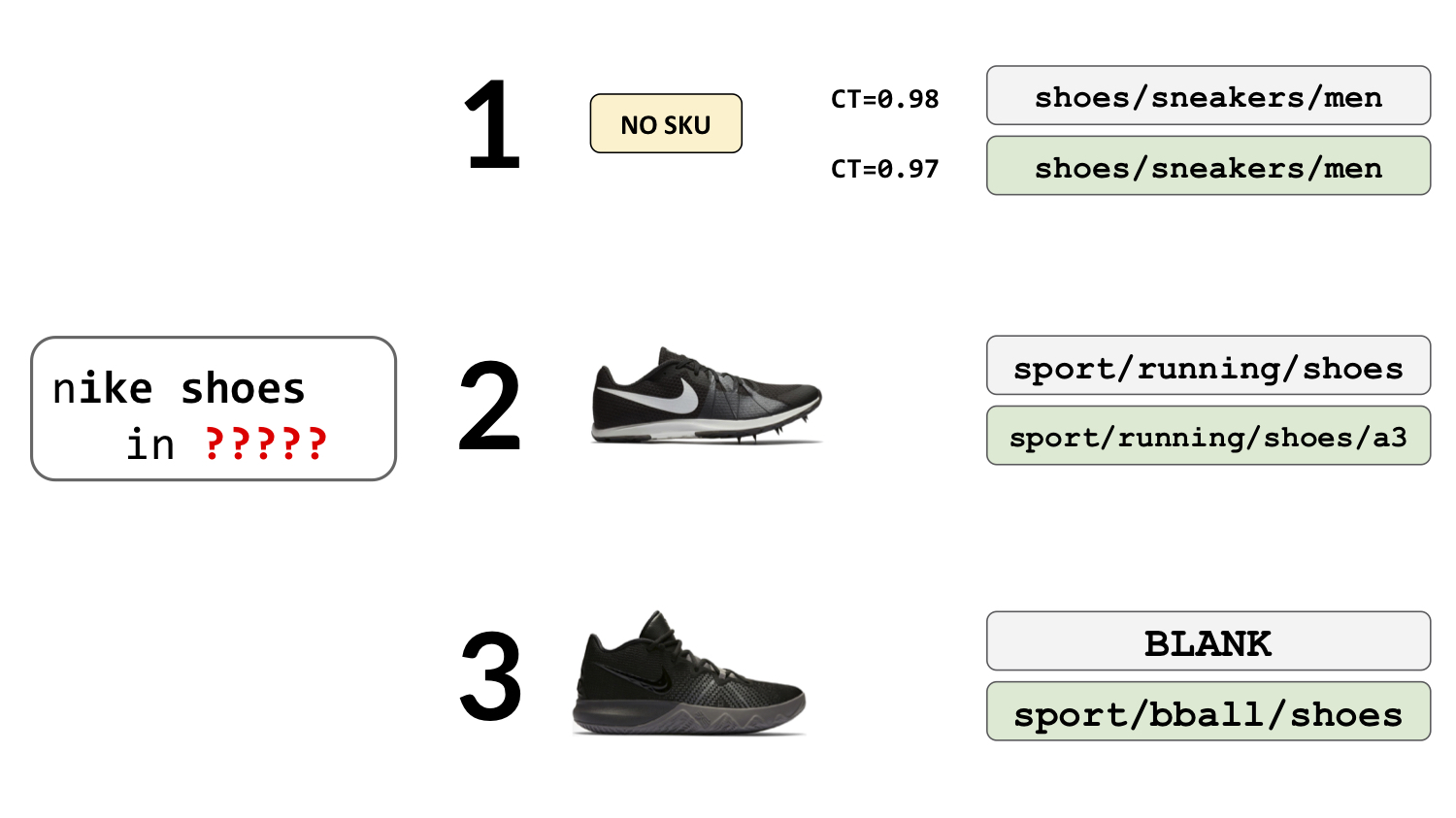}
\end{center}
\caption{Sample \textbf{SessionPath} predictions for the candidate query ``nike shoes'', with two thresholds (gray, green) and three sessions, $1,2,3$ (no product for session $1$, a pair of running shoes for $2$, a pair of basketball shoes for $3$). The model reacts quickly both across sessions (switching to relevant parts of the underlying product catalog) and across threshold values, making more aggressive decisions at a lower value (green). }
\label{fig:predictions}
\end{figure}

\textit{This} research paper introduced \textbf{SessionPath}, a personalized and scalable model that dynamically suggests product paths in type-ahead systems;~\textbf{SessionPath} was benchmarked on data from two shops and tested against count-based and neural models, with explicit complexity-accuracy trade-offs. Finally, we proposed a confidence-based decision rule inspired by customer discussions: by abstracting away model behavior in one parameter, we wish to solve the often hard interplay between business requirements and machine behavior; furthermore, by leveraging a hierarchical structure of product concepts, the model produces predictions that are suitable to a \textit{prima facie} human inspection (e.g. Figure~\ref{fig:predictions}).

While our evaluation shows very encouraging results, the next step will be to A/B test the proposed models on a variety of target clients:~\textbf{Shop 1} and \textbf{Shop 2} data comes from search logs of a last-generation search engine, which possibly skewed model behavior in subtle ways. With more data, it will be possible to extend the current work in some important directions: 

\begin{enumerate}
    \item while \textit{this} work showed that \textbf{SessionPath} is effective, the underlying deep architecture can be improved further: on one hand, by doing more extensive optimization; on the other, by focusing on how to best perform linguistic generalization: \textit{transfer learning} (between tasks as proposed by \citet{Skinner2019EcommerceQC}, or across clients, as described in \citet{CoveoECNLP20}) is a powerful tool that could be used to improve performances further;
    \item the same model can be applied with almost no changes to the search workflow, to provide a principled way to do personalized query scoping. A preliminary A/B test on~\textbf{Shop 1} using the \textit{MLP} model on a minor catalog facet yielded a small (2\%) but statistically significant improvement ($p<0.05$) on click-through rate and we look forward to extending our testing;
    \item we could model path depth~\textit{within the decoder itself}, by teaching the model when to stop; as an alternative to learning a decision rule in a supervised setting, we could leverage reinforcement learning and let the system improve through iterations - in particular, the choice of cutting the path for a given query and session vector could be cast in terms of contextual bandits;
    \item finally,~\textit{precision} and~\textit{recall} at different depths are just a first start; preliminary tests with \textit{balanced accuracy} on selected examples show promising results, but we look forward to performing user studies to deepen our understanding of the ideal decision mechanism.
\end{enumerate}

Personalization engines for digital shops are expected to drive an increase in profits by 15\% by the end of 2020~\citep{Gartner2018}; facet suggestions help personalizing the search experience as early as in the type-ahead drop-down window: considering that search users account globally for almost 14\% of the total revenues~\citep{Charlton2013}, and that category suggestions may improve click-through-rate and reduce cognitive load, \textbf{SessionPath} (and similar models) may play an important role in next-generation online experiences.

\section*{Acknowledgments}
Thanks to (in order of appearance) Andrea Polonioli, Federico Bianchi, Ciro Greco, Piero Molino for helpful comments to previous versions of this article. We also wish to thank our anonymous reviewers, who greatly helped in improving the clarity of our exposition.

\bibliography{acl2020}
\bibliographystyle{acl_natbib}

\appendix

\section{Architectural Considerations}
\label{sec:architecture}
\begin{figure}[H]
\includegraphics[width=7.5cm]{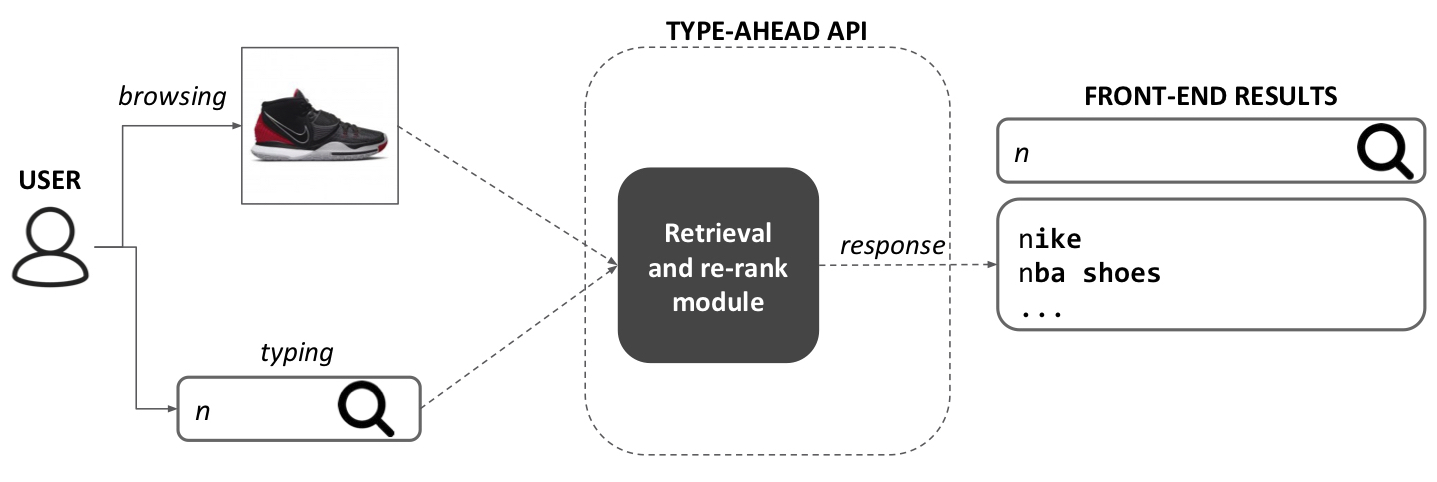}
\caption{High-level functional overview of an industry standard API for type-ahead suggestions: query seed and possibly session information about the user are sent by the client to the server, where some retrieval and re-ranking module produces the final top-\textit{k} suggestions and prepares the response for front-end consumption.}
\label{fig:oldarch}
\end{figure}

Figure~\ref{fig:oldarch} represents a functional overview of a type-ahead service: when \textit{User X} on a shop starts typing a query after browsing some products, the query seed and the session context are sent to the server. An existing engine - traditional or neural - will then take the query and the context and produce a list of top-\textit{k} query candidates, ranked by relevance, which are then sent back to the client to populate the dropdown window of the search bar. 

\begin{figure}
\begin{center}
    \includegraphics[width=7.0cm]{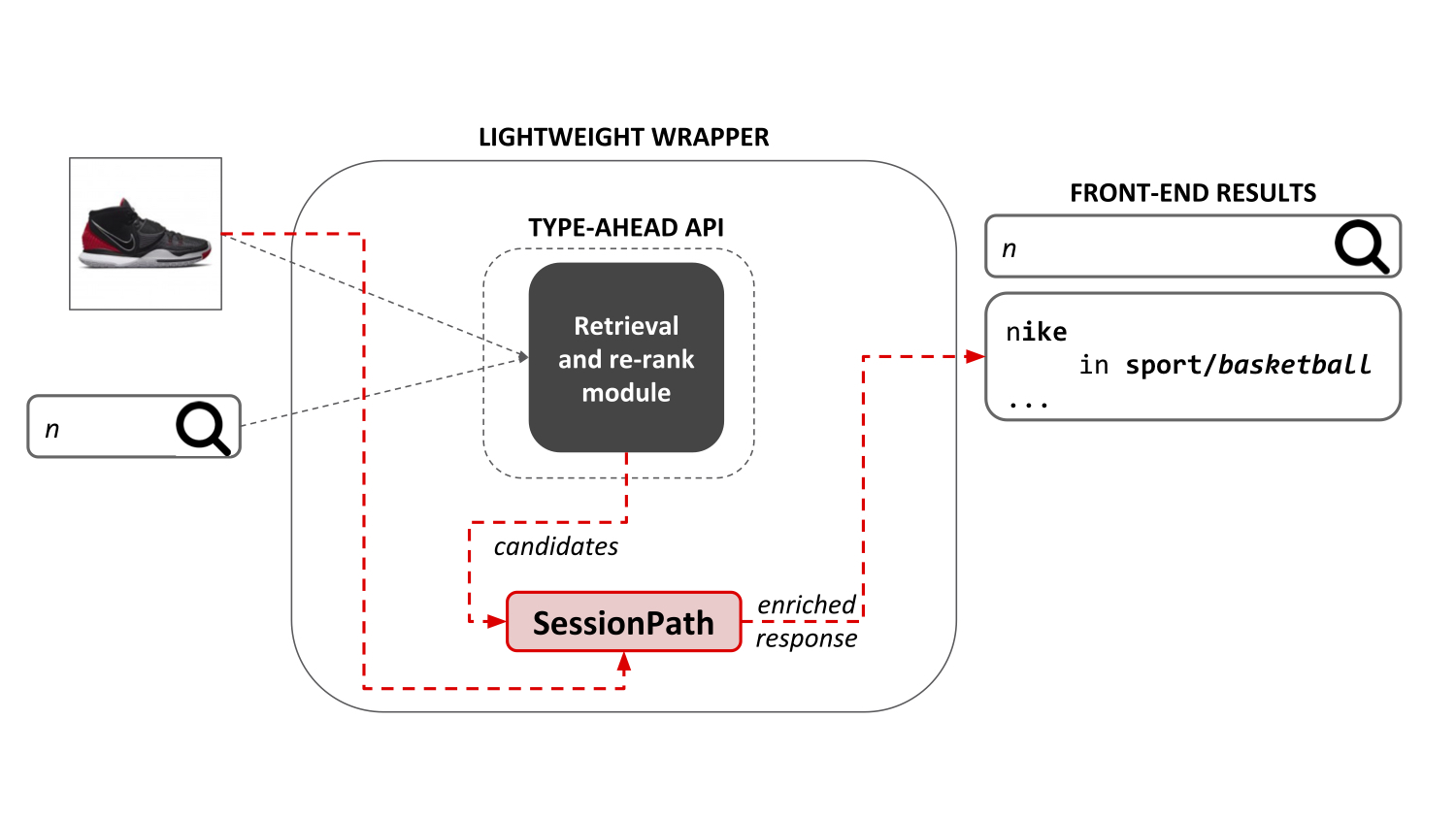}
\end{center}
\caption{A lightweight \textbf{SessionPath} functional integration: starting from a standard flow (Figure~\ref{fig:oldarch}, a simple wrapper around the existing module sends the same session information and the top-\textit{n} suggestions to \textbf{SessionPath}, for dynamic path prediction. The final response is then obtained by simply augmenting the existing response containing query candidates with category predictions.}
\label{fig:newarch}
\end{figure}

As depicted in Figure~\ref{fig:newarch}, category suggestions can be quickly added to any existing infrastructure by treating the current engine as a ``black-box'' and adding path predictions at run-time for the first (or the first \textit{k}, since requests to the model at that point can be batched with little overhead) query candidate(s). In this scenario, the decoupling between retrieval and suggestions is absolute, which may be a good idea when the stacks are very different (say, traditional retrieval \textit{and} neural suggestions), but less extreme solutions are obviously possible. The crucial engineering point is that path prediction (using any of the methods from Section~\ref{sec:experiments}) can be added and tested quickly, with few conceptual and engineering dependencies: the more traditional the existing stack, the more an incremental approach is recommended: count-based first - since predictions can be served simply from an in-memory map -, \textbf{MLP} second - since predictions require a small neural network, but they are fast enough to only require CPU at query time -, and finally the full \textbf{SessionPath} - which requires dedicated hardware considerations to be effective in the time constraints of the type-ahead use case. As a practical suggestion, we also found quite effective when using simpler models (e.g.~\textbf{MLP}) to first test it at a \textit{given depth}: for example, you start by only classifying the most likely nodes in template~\textit{sport / ?}, and then incrementally increase the target classes by adding more diverse paths.

Adding a lightweight wrapper around the original bare-bone endpoint allows for other improvements as well: for example, considering typical power-law of query logs, a caching layer can be used to avoid a full retrieving-and-rerank iteration for frequent queries; obviously, this and similar features are independent from \textbf{SessionPath} itself.

\section{Metrics Calculation: a Worked-Out Example}
\label{sec:metricexample}
\begin{figure}
\begin{center}
    \includegraphics[width=7.5cm]{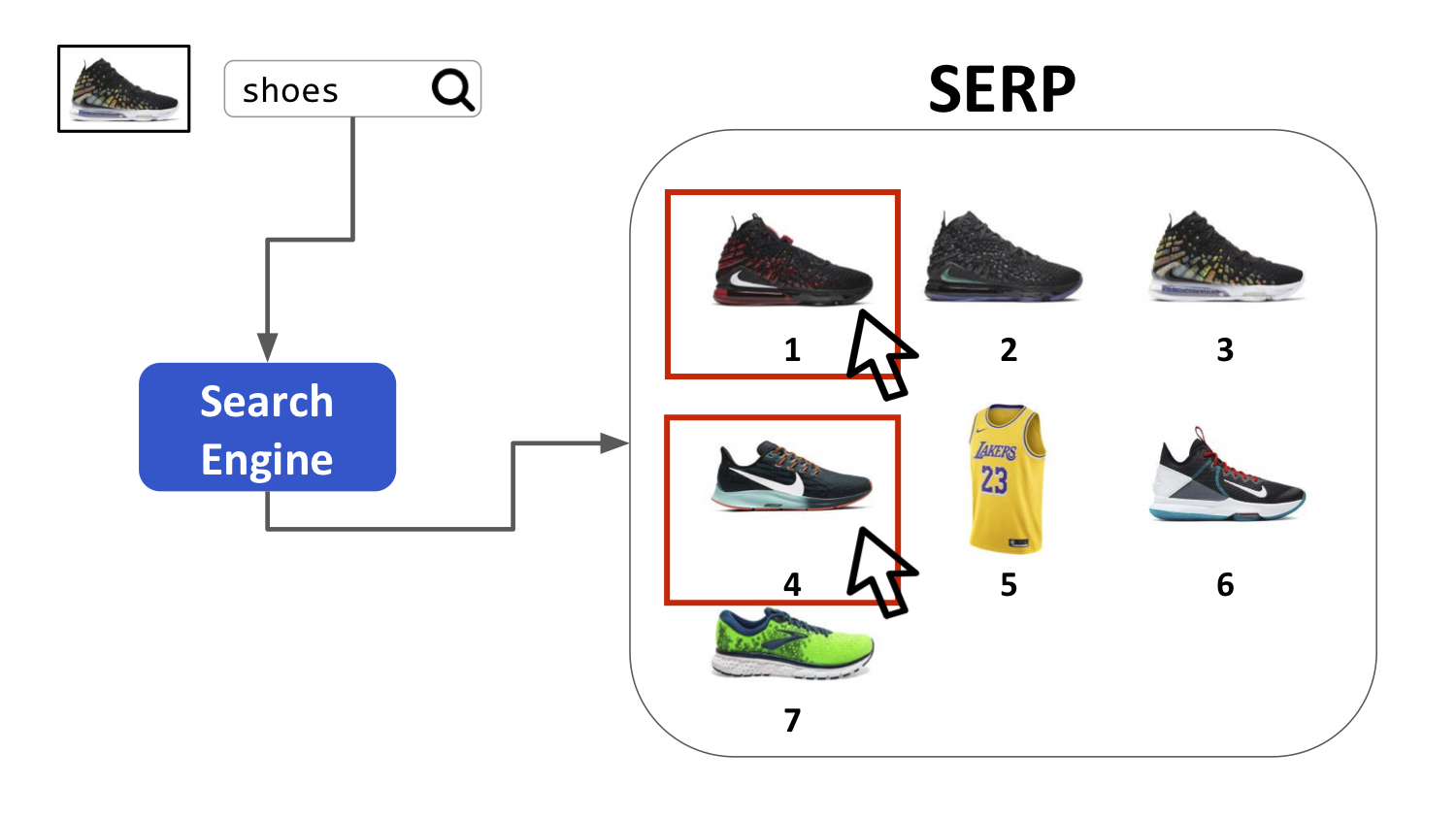}
\end{center}
\caption{A sample row in the test set, displaying search results ($7$ products in $4$ paths) for the query ``shoes'' and a session containing a pair of \textit{LeBron James} basketball shoes. In this example, the shopper clicked on products $P_1$ and $P_4$.}
\label{fig:metric}
\end{figure}

For the sake of reproducibility, we present a worked out example of metrics calculations for offline testing of the decision module (Section~\ref{sec:decision}). Figure~\ref{fig:metric} depicts an historical interaction from the search logs: a session containing a product, a query issued by the user and the search result page (``serp''), containing seven items belonging to the following paths:
\bigbreak
$P_1=$~\textit{sport / basketball / lebron}

$P_2=$~\textit{sport / basketball / lebron}

$P_3=$~\textit{sport / basketball / lebron} 

$P_4=$~\textit{sport / running / sneakers}

$P_5=$~\textit{sport / basketball / jerseys} 

$P_6=$~\textit{sport / basketball / curry}

$P_7=$~\textit{sport / running / sneakers}.
\bigbreak
Click-through data (i.e. products in the serp clicked by the user) indicates that $P_1$ and $P_4$ are relevant, and so the associated paths are ground truths (\textit{sport/basketball/lebron} and~\textit{sport/running/sneakers}). We now present the full calculations in three scenarios, corresponding to three level of depths in the predicted path.
\bigbreak
\textbf{Scenario 1 (general)}: prediction is \textit{sport}. In this case, result set would be intact, so: \textit{True Positives} (\textbf{TP}) are $P_1,P_2,P_3,P_4,P_7$, \textit{False Positives} (\textbf{FP}) are $P_5,P_6$, \textit{False Negatives} (\textbf{FN}) are $\emptyset$.~\textit{Precision} is: \textbf{TP} / (\textbf{TP} + \textbf{FP}) = $5 / (5 + 2) = 0.71$,~\textit{Recall} is: \textbf{TP} / (\textbf{TP} + \textbf{FN}) = $5 / (5 + 0) = 1.0$ (with no cut, all truths are retrieved so $1.0$ is the expected result).
\bigbreak
\textbf{Scenario 2 (intermediate)}: prediction is \textit{sport/basketball}. In this case, filtering the result set according to the decision made by the model would give $P_1,P_2,P_3,P_5,P_6$ as the final set. So:~\textbf{TP}~=~$P_1,P_2,P_3$,~\textbf{FP}~=~$P_5,P_6$,~\textbf{FN}~=~$P_4,P_7$;~\textit{Precision}~=~$3 / (3 + 2) = 0.6$,~\textit{Recall}~=~$3 / (3 + 2) = 0.6$.
\bigbreak
\textbf{Scenario 3 (specific)}: prediction is~\textit{sport / basketball / lebron}. In this case, filtering the result set according to the decision made by the model would give $P_1,P_2,P_3$ as the final set. So: \textbf{TP}~=~$P_1,P_2,P_3$,~\textbf{FP}~=~$\emptyset$,~\textbf{FN}~=~$P_4,P_7$;~\textit{Precision}~=~$3 / (3 + 0) = 1.0$,~\textit{Recall}~=~$3 / (3 + 2) = 0.6$.
\bigbreak
The full calculations show very clearly the natural trade-off discussed at length in Section~\ref{sec:decision}: the deeper the path, the more precise are the results but also the higher the chance of hiding valuable products from the shopper.

\end{document}